\documentclass[preprint,
3p,
]{elsarticle}

\usepackage[utf8]{inputenc}
\usepackage[T1]{fontenc} 
\usepackage[english]{babel}
\usepackage{autobreak}

\usepackage{amsmath, amssymb, amsthm}
\usepackage{mathtools, mathrsfs}

\usepackage[colorlinks=true, citecolor=blue, linkcolor=blue, urlcolor=blue]{hyperref}

\usepackage{graphicx}
\usepackage{caption}
\DeclareCaptionFont{myfont}{\fontsize{10pt}{12pt}\selectfont}
\usepackage{subcaption}

\usepackage{setspace,tabularx}
\usepackage{multirow}
\usepackage{makecell}
\usepackage{booktabs, array} 
\newcolumntype{Y}{>{\arraybackslash}X}
\newcolumntype{P}[1]{>{\centering\arraybackslash}m{#1}}
\newcolumntype{J}[1]{>{\arraybackslash}m{#1}}

\newdefinition{defi}{Definition}

\newcommand{\ie}{i.e.}
\newcommand{\R}{\mathbb{R}}

\begin{document}

\begin{frontmatter}

\title{Inference of Unknown Dynamical Components Using Next Generation Reservoir Computing: From Chaotic Systems to Climate Data}

\author[1]{Jule Budnick}
\ead{125136633@umail.ucc.ie}
\author[1]{Andrew Keane}
\ead{andrew.keane@ucc.ie}
\author[1,2]{Serhiy Yanchuk\corref{cor1}}
\ead{syanchuk@ucc.ie}

\cortext[cor1]{Corresponding author}
\affiliation[1]{organization={School of Mathematical Sciences, University College Cork},
city={Cork},
country={Ireland}}
\affiliation[2]{organization={Potsdam Institute for Climate Impact Research},
city={Potsdam},
country={Germany}}

\begin{abstract}
We investigate next generation reservoir computing (NGRC) as a data-driven approach for inferring unseen components of dynamical systems. We compare NGRC with traditional reservoir computing (RC) using the Lorenz and Rössler system, where two unknown components are inferred from one given component. For both systems, NGRC achieves accurate results while requiring fewer training data and less computational time than RC. We identified an inverse proportional behavior between the number of time-delayed steps needed for NGRC and the temporal resolution, indicating that the physical time span covered by the delay interval is an important factor in determining the required number of delayed steps. Finally, we apply NGRC to the observational climate data of ENSO (El Niño--Southern Oscillation) and infer one observable from the remaining variables. Despite the noise and complexity of the real-world data, the NGRC shows promising results. Our findings demonstrate the potential of NGRC for efficient inference of unseen components in both controlled dynamical systems and real-world data.
\end{abstract}

\begin{keyword}
NGRC \sep RC \sep Unseen components \sep Chaotic dynamical systems \sep ENSO
\end{keyword}

\end{frontmatter}

\section{Introduction}

Data-driven approaches, especially machine learning (ML) algorithms, have become the state-of-the-art framework to predict and reproduce chaotic motions of dynamical systems in their various application areas \cite{introduction_ds_ml,introduction_ds_ml_1,ds_ml_climate,ds_ml_finance,ds_ml_finance_1}. 
A special ML paradigm is \textit{reservoir computing} (RC), whose time-series data may be successfully analyzed through the lens of dynamical systems theory \cite{introduction_reservoir,lsm,yan2024emerging}. 
A reservoir computer is a recurrent neural network (RNN) characterized by interconnected neurons that are arranged in a so-called \textit{reservoir} (a pool of nodes) instead of layers. 
In contrast to standard RNN algorithms, only the output weights of the RC are trained with a simple and efficient least squares method. 
All internal network parameters, such as the strengths of the node-to-node connections or the input coefficients, are chosen randomly. With these simplifications, the RC algorithm enables shorter training times with smaller training data sets, compared to standard RNNs, even for high-dimensional time-dependent tasks \cite{Vlachas2020}. Despite its simplicity, the RC performs as well as other ML techniques for certain tasks \cite{rc_bompas}. 
It is widely used for predicting dynamical systems output \cite{esn,rc_griffith}, learning climate data \cite{rc_lu,rc_roehm}, and inferring unseen variables \cite{rc_lu2}. 
Furthermore, a class of RCs adapted to nonlinear systems with delayed feedback, known as delay-based RCs, have shown excellent performance with efficient and fast computation \cite{ds_delay,VanderSande2017,Larger2017,Hart2019,Stelzer2019,Goldmann2020,yanchuk}. 

RC is closely related to the vector autoregressive (VAR) method for modeling and forecasting multiple interrelated time series. In the specific case of a linear RC (\ie, the activation function is linear), \cite{rc_bolt} shows the equivalence of a RC with linear readout to a VAR as well as of an quadratic readout reservoir to a nonlinear VAR. This may explain the surprising success and efficiency of RCs \cite{Jaurigue2022}. Further work has also shown that under certain conditions a RC is a universal approximator \cite{univ_approx}. 

When implementing the RC approach an optimization of \textit{hyperparameters} is necessary. These hyperparameters characterize the general properties of a RC and strongly influence its performance. However, the corresponding parameter space can be large, including the number of reservoir nodes, the spectral radius of the network, the sparsity or the input amplitude \cite{Lukoeviius2012}. There are some explicitly known results, for example, in relation to the spectral radius of the network and the need for recurrent connections \cite{introduction_reservoir,Lukoeviius2012}. Additionally, there are more general results regarding the sparsity or low-connectivity of the reservoir \cite{rc_griffith,Song2010}, or the leaking rate \cite{Lukoeviius2012,gradient_descent}. Nonetheless, in practice, an extensive parameter search is still required. Possible optimization algorithms are a grid search \cite{grid_search}, Bayesian \cite{bayesian_1,bayesian_2} or gradient descent methods \cite{gradient_descent}, all of which are computationally expensive or only work on a continuous parameter set.

As a further development of RC, \textit{next generation reservoir computing} (NGRC) \cite{ngrc} greatly simplifies the optimization procedure. The NGRC algorithm depends on only two hyperparameters: the number of time delayed steps and a regression parameter. At each time step $i$, the network receives the input data $X_j$ from the previous $k$ time (delayed) steps ($i\geq j >i-k$). These delayed inputs constitute the feature vector, so the choice of $k$ determines both its size and the associated computational expense. The regression parameter results from a least squares method used to train the algorithm. Recent studies have demonstrated the advantages of NGRC over
traditional RC when only limited training data are available \cite{raeth}.

According to \cite{ngrc}, there are three common benchmark tasks which a ML algorithm should accomplish for a dynamical system: (i) short-term forecasting, (ii) reconstructing the attractor (long-term forecasting), and (iii) inferring unseen data. 
We investigate the performance of NGRC and compare it to the standard RC for the \textit{inference of unknown dynamical components} of chaotic systems and climate data. 

Inference tasks provide a means to recover unobserved variables of a dynamical system from a subset of variables that can be measured. This can be achieved if a dynamical system exhibits sufficiently strong internal couplings between variables and the observed variables carry enough information to reconstruct the unseen components. Mathematical and statistical techniques including state‑space reconstruction or projections into latent spaces using autoencoders or proper orthogonal decomposition, allow applications such as inferring neural activity from partial electrode recordings \cite{talukder2022deep} or reconstructing surface pressure fields from sparse measurements \cite{qiu2025compressed}. The same logic is highly valuable in climate science, where observational constraints vary widely across variables. Identifying which climate variables can be reliably inferred from more accessible measurements reduces the burden on observing networks and increases resilience to missing or intermittent data \cite{chao2025learning,kadow2020artificial,konrad2017reduced}. 

In this paper we investigate how the NGRC performs against the traditional RC for the task of inferring unseen variables/observables. We consider the Lorenz \cite{Lorenz1963} and Rössler \cite{roessler_or} chaotic systems, as well as climate data measurements related to the El Ni\~no--Southern Oscillations (ENSO) system \cite{ENSO_or,ENSO_overview}. 
Table~\ref{tab:tasks} gives an overview of some known applications of NGRC and RC to these three systems, as well as new contributions from this study. 

\begin{table}[!ht]
	\centering
    {\fontsize{10pt}{12pt}\selectfont
	\begin{tabularx}{\textwidth}{ P{2.2cm} P{6cm} P{5.1cm} } 
		\toprule[1.5pt]
		& \textbf{Previous results} & \textbf{This study}\\
		\midrule[1.5pt]
		\\[-.6cm]

		\multirow{4}{*}{\makecell{NGRC}} & Inference of one component for Lorenz system \cite{ngrc} & Inference of two components for Lorenz system\\  
		& -- & Inference of two components for Rössler system\\
        & -- & Inference of one component for ENSO\\ \hline
		\multirow{6}{*}{\makecell{RC}} & Inference of two components for Lorenz system \cite{rc_lu2} & Comparing NGRC to RC for Lorenz system \\ 
		&Inference of two components for Rössler system \cite{rc_lu2}& Comparing NGRC to RC for Rössler system\\
        & \multirow{2}{*}{--}& Inference of one component for ENSO\\
        & & Comparing NGRC to RC for ENSO
	\end{tabularx}
	\caption{\label{tab:tasks}A summary of known previous results on inference of unknown components and new results from this work.}}
\end{table}

In the following two sections we provide a brief summary of the NGRC methodology and the three dynamical systems we consider. In Section~\ref{sec:results} we demonstrate the applicability of NGRC to the inference of unknown components of chaotic dynamical systems in the previously untested setups: two unknown components of the Lorenz system, two unknown components of the Rössler system, and one unknown data series from ENSO measurements. The latter shows the usefulness of NGRC for complex and noisy real-world data. We find that neither algorithm consistently outperforms the other regarding different metrics, however, the NGRC has the advantage of requiring less training data and having a significant reduced computational cost. Both approaches perform surprisingly well with the noisy, real-world ENSO data by capturing prominent trends. However, it is also clear that there is much room for improvement. Additionally, we demonstrate how the optimal choice of the number of time delayed steps depends on the temporal resolution of the data and the desired accuracy of inference.

\section{Next Generation Reservoir Computing}

The architecture of a traditional RC is described in \cite{lsm} or \cite{esn}. The NGRC framework was first described by \cite{ngrc} and we now present the main ideas of their approach. The first step is the construction of a \textit{feature vector} directly from the input data.  

\begin{defi}(\textbf{Feature vector})\label{def:feature_vector}
	For a time discretization $t_i$, $i=1,..,n$, let $X_i \in \R^{N_\text{in}}$ be the input data at time $t_i$. The \textit{feature vector} of the NGRC is given by
	\begin{align}
		\label{def:ototal}
		\mathbb{O}_{\text{total},i} \coloneqq c \oplus \mathbb{O}_{\text{lin}, i} \oplus \mathbb{O}_{\text{nonlin}, i}^{(p)} \in \R^{N_\text{total}},
	\end{align}
	where $c \in \R$ is a constant, $\mathbb{O}_{\text{lin}, i}$ represents the linear and $\mathbb{O}_{\text{nonlin}, i}^{(p)}$ the nonlinear part, and $\oplus$ denotes vector concatenation. The linear part of the feature vector is given by
	\begin{align*}
	\mathbb{O}_{\text{lin},i} \coloneqq X_i \oplus X_{i-1} \oplus \dots \oplus X_{i-(k-1)} \in \R^{kN_\text{in}},
	\end{align*} 
	where $k<i$ is the number of time delayed steps for which the data $X_j$, $i-k<j\leq i$ is used. The nonlinear part of the feature vector is a function of the linear part defined as
	\begin{align*}
	\mathbb{O}_{\text{nonlin},i}^{(p)} \coloneq \mathbb{O}_{\text{lin},i}\ \lceil \otimes \rceil\  \mathbb{O}_{\text{lin},i}\ \lceil \otimes \rceil \dots \lceil \otimes \rceil\ \mathbb{O}_{\text{lin},i},
	\end{align*}
	where the operator $\lceil \otimes \rceil$ denotes the calculation of the outer product, then takes the upper triangular matrix (the unique monomials) and flattens them into a vector. The product is taken $p$-times, hence, the parameter $p$ describes the maximal polynomial power appearing in the feature vector. The total dimension $N_\text{total}$ depends on the input dimension $N_\text{in}$, the time delayed steps $k$, the maximal power $p$ and is equal to $\binom{N_{in}k + p}{p}$.
\end{defi}

For example, the feature vector for a system with $N_{in}=3$, described by $X=[x,y,z]$, at time step $t_i$ with $k=2$ and $p=2$ is given by
\begin{align*}
\mathbb{O}_{\text{total},i}=
[& c_i,x_{i},y_{i},z_{i},x_{i}^2,x_{i}y_{i},x_{i}z_{i},y_{i}^2,y_{i}z_{i},z_{i}^2,\dots \\
& x_{i-1},y_{i-1},z_{i-1},x_{i-1}^2,x_{i-1}y_{i-1},x_{i-1}z_{i-1},y_{i-1}^2,y_{i-1}z_{i-1},z_{i-1}^2, \dots, \\
& x_{i}x_{i-1},x_{i}y_{i-1},x_{i}z_{i-1},
y_{i}x_{i-1},y_{i}y_{i-1},y_{i}z_{i-1},
z_{i}x_{i-1},z_{i}y_{i-1},z_{i}z_{i-1}].
\end{align*}
The size of the feature vector will grow very quickly with larger choices of $k$ and $p$. \cite{ngrc} has shown that feature vectors containing only low-order monomials may be sufficient to perform accurate predictions. Therefore, we choose $p=2$ for this study.
 
The training of a NGRC focuses on the accurate prediction from the state $X_i$ to $X_{i+1}$. This is described in the following definition.

\begin{defi}(\textbf{NGRC setup and training})
	Let $X_i \in \R^{N_\text{in}}$ be the input data at time $t_i$ and let $\mathbb{O}_{\text{total},i}\left( X_i, X_{i-1}, \dots, X_{i-(k-1)}\right) \in \R^{N_\text{total}}$ the feature vector according to Definition \ref{def:feature_vector}. The NGRC model is given by
	\begin{align}
		\label{eq:ngrc}
		X_{i+1} = X_i + W_\text{out} \mathbb{O}_{\text{total},i} \left( X_i, X_{i-1}, \dots, X_{i-(k-1)}\right) \quad \text{for } i=1,..,n,
	\end{align}
	where the weight matrix $W_\text{out}$ is trained analogously to the RC approach. In matrix notation, Equation \eqref{eq:ngrc} leads to
	\begin{align}
		\Delta X = W_\text{out} \mathbb{O}_\text{total},
	\end{align}
	where $\Delta X = \left(X_2 - X_1, \dots, X_{n+1} - X_n\right) \in \mathbb{R}^{N_\text{in}\times n}$, $W_\text{out} \in \R^{N_\text{in} \times N_\text{total}}$ and $\mathbb{O}_\text{total} = (\mathbb{O}_{\text{total},1}, \dots, \mathbb{O}_{\text{total},n}) \in \R^{N_\text{total}\times n}$.
	Therefore, Tikhonov regularization (also known as ridge regression) can be applied to find $W_{\text{out}}$
	\begin{align}
		W_{\text{out}} &=
     \operatorname{\text{argmin}}_{\widetilde{W}_{\text{out}}} 
     \left(
         \| \Delta X_d - \widetilde{W}_\text{out} \mathbb{O}_\text{total} \|_F^2 + 
         \alpha \| \widetilde{W}_\text{out} \|_F^2 
     \right) \nonumber \\
       &= \Delta X_d \mathbb{O}_\text{total}^T\left(\mathbb{O}_\text{total} \mathbb{O}_\text{total}^T + \alpha \mathbb{I}\right)^{-1},
	\end{align}
	where $\Delta X_d \in \mathbb{R}^{N_\text{in}\times n}$ represents the desired difference between two states of the system at consecutive time steps (given by the training data set), $\alpha$ is the regularization parameter and
    $$
    \| A \|_F \coloneq \sqrt{\sum_{i=1}^n \sum_{j=1}^m a_{ij}} \quad \text{for } A \in \R^{m\times n}
    $$
    the Frobenius-Norm of a matrix.
\end{defi}

For the inference of unknown components, an adjustment of the NGRC algorithm is necessary. Consider a $N_\text{sys}$-dimensional system. We now want to predict $N_\text{infer}$ unknown components from the other $N_\text{in}$ given components, such that $N_\text{sys} = N_\text{in}+N_\text{infer}$. Therefore, the flow equation \eqref{eq:ngrc} needs the following modification:
\begin{align*}
	Y_{i+1} = W_\text{out} \mathbb{O}_{\text{total},i},
\end{align*}
where $Y_{i+1} \in \mathbb{R}^{N_\text{infer}}$ represents the vector of the $N_\text{infer}$ components to predict. For $p=2$, the nonlinear part of the feature vector at time $t_i$ is given by
\begin{align*}
    \mathbb{O}_{\text{nonlin},i}^{(p)}  = \mathbb{O}_{\text{nonlin},i}^{(2)} &= \mathbb{O}_{\text{lin},i}\ \lceil \otimes \rceil\  \mathbb{O}_{\text{lin},i}.
\end{align*}
Hence, the outer product is a symmetric $(kN_\text{in}\times kN_\text{in})$ matrix and the dimension of the nonlinear part is $\frac{kN_\text{in} (kN_\text{in}+1)}{2}$. Therefore, $\mathbb{O}_{\text{total},i}$ has
\begin{align*}
    1 + kN_\text{in} + \frac{kN_\text{in}(kN_\text{in}+1)}{2}
\end{align*}
components.

It is useful to mention the following two main advantages of NGRC over the RC, as discussed in \cite{ngrc}. First, the randomly chosen matrices in the RC approach are replaced by a fixed construction of the feature vectors. Therefore, the number of possible hyperparameters is reduced. This significantly reduces the number of required training data points.
Secondly, there are fewer warm-up points (\ie, the points which are necessary to initialize the algorithm) and training points needed in comparison to RC. For a traditional RC, the warm-up period can last from $10^3$ to $10^5$ data points \cite{rc_roehm,rc_griffith,rc_lu} since longer warm-up times are needed to guarantee that the RC becomes independent of the RC initial conditions. The necessary number of warm-up points in the NGRC algorithm is merely the number of time delayed steps $k$ which are needed to create the feature vector at the first time step.

\section{Three systems used in this study}

The three systems considered are the Lorenz system, the Rössler system and as a climate data application the El Ni\~no--Southern Oscillation (ENSO). The Lorenz and Rössler systems are three-dimensional models that exhibit chaos and contain a strange attractor for a suitable choice of parameters. The datasets used to train the NGRCs are generated by numerical integration of these models using the fourth-order Runge-Kutta method. The application to the ENSO phenomenon, on the other hand, is based on observational data only. In the following we describe each system.

\subsection{Lorenz system}
\label{sec:theorylorenz}

\begin{figure}[!ht]
	\centering
	\includegraphics[width=\textwidth]{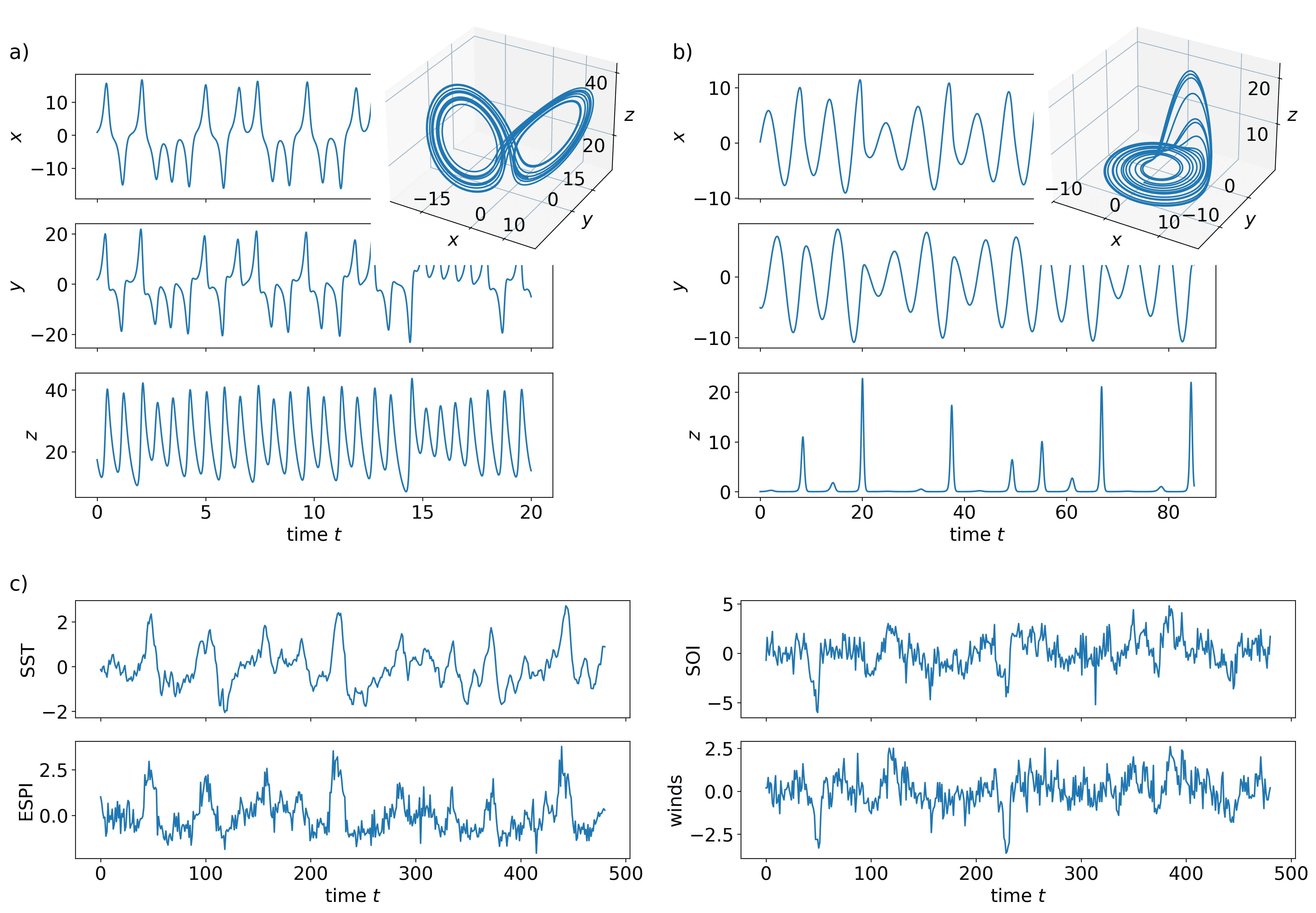}
	\caption{\label{fig:timeseries}a) A chaotic trajectory of the Lorenz system for $\sigma = 10$, $b = \frac{8}{3}$, and $r = 28$. b) A chaotic trajectory of the Rössler system \eqref{eq:roessler} for $a=0.2,\ b=0.2$ and $c=5.7$. c) ENSO data sets plotted against time (in months) for sea surface temperature (SST), southern oscillation index (SOI), precipitation index (ESPI) and zonal winds.}
\end{figure}

The fundamental work on chaotic behavior goes back to Edward Lorenz who developed a system to model a simplification of atmospheric convection \cite{Lorenz1963}. This rather simple system exhibits chaos and contains a strange attractor. The three-dimensional Lorenz system is given by
\begin{equation}
    \begin{aligned}
	\dot{x}(t) &= \sigma \left(y(t)-x(t)\right),\\
	\dot{y}(t) &= rx(t) - y(t) -x(t)z(t),\\
	\dot{z}(t) &= x(t)y(t) - bz(t),
    \end{aligned}
    \label{eq:lorenz}
\end{equation}
with $\left(x(t),y(t),z(t)\right) \in \R^3$ and parameters $\sigma, r, b \in \R$ \cite{Lorenz1963}. We will consider the original parameter choice of Lorenz: $\sigma = 10$, $b = \frac{8}{3}$, and $r = 28$, for which the system exhibits a strange attractor \cite{williams1979structure,tucker1999lorenz}. Fig.~\ref{fig:timeseries}(a) shows a trajectory of the Lorenz system, in terms of time $t$ and the phase space, calculated by numerically integrating Eqs.~(\ref{eq:lorenz}) and removing transients.

There are already several studies of short- and long-term forecasting for the Lorenz system. For example, a forecasting task has been performed by \cite{rc_bompas} or \cite{rc_griffith} for the traditional RC. Similarly, \cite{ngrc} showed forecasting of the Lorenz attractor using NGRC and achieved accurate short-term predictions and nearly exact reconstruction of the attractor. For the inference of unknown components, \cite{rc_lu2} showed accurate predictions of two unseen components with RC. For the NGRC algorithm, \cite{ngrc} performed the inference of the $z$ component from both the $x$ and $y$ components. In this work, we perform an inference of two unknown components using NGRC. 

\subsection{Rössler system}
\label{sec:theoryreossler}

The Rössler system is given by
\begin{equation}
\begin{aligned}
	\dot{x}(t) &= - \left(y(t)+z(t)\right),\\
	\dot{y}(t) &= x(t) + ay(t),\\
	\dot{z}(t) &= b+z(t)\left(x(t)-c\right),
    \end{aligned}
    \label{eq:roessler}
\end{equation}
with $\left(x(t), y(t), z(t)\right) \in \R^3$ and $a,b,c \in \R$. We consider the parameter values $a=0.2,\ b=0.2$ and $c=5.7$ as in \cite{roessler_or}. Fig.~\ref{fig:timeseries}(b) shows an example of a chaotic trajectory as time series and also in the $(x,y,z)$-space. For the Rössler system, \cite{rc_bompas} demonstrates long- and short-term forecasting of a chaotic trajectory. The inference of two unknown components was performed in \cite{rc_lu2} using RC. However, no application of the NGRC to the inference of unknown components exists for the Rössler system \eqref{eq:roessler}.

\subsection{El Niño–Southern Oscillation data series}
\label{sec:theoryenso}

The El Ni\~no--Southern Oscillation (ENSO) describes a climate phenomenon in the equatorial Pacific. ENSO is one of the most important ``weather-makers'' with worldwide changes in temperature and rainfall \cite{ENSO_overview}. An extreme ENSO can affect agriculture \cite{ENSO_agri}, ecosystems \cite{ENSO_echosys_1,ENSO_echosys_2}, power generation \cite{ENSO_powergen} and extreme weather events worldwide such as wildfires, floods and tropical cyclones \cite{ENSO_general,ENSO_extreme_1,ENSO_extreme_2}. Therefore, considering the massive impact of the ENSO on the global climate and economy, an ability to understand and precisely forecast its pattern is crucial.

ENSO constitutes two interacting components in the ocean and atmosphere of the equatorial Pacific. \textit{El Ni\~no} events are associated with large sea-surface temperatures (relative to the long-term mean) and the \textit{Southern Oscillations} that describe fluctuations in the surface air pressure. Additionally, other weather components play a role, such as winds and precipitation \cite{ENSO_overview}.

\begin{table}[!htp]
	\centering
	\begin{tabularx}{\textwidth}{P{1.3cm} J{\dimexpr\linewidth-9.8cm} P{2cm} P{2.2cm} P{2.2cm}}
		\toprule[1.5pt]
		\textbf{Name} & \centering \textbf{Description} & \textbf{Time Coverage} & \textbf{Dataset} & \textbf{Source}\\ 
		\midrule[1.5pt]
		SST & Sea surface temperature anomalies in the Ni\~no 3.4 region (5°N to 5°S; 170°W to 120°W) \cite{Reynolds2002} & monthly, since 1950  & HadISST1.1 & NOAA CPC\\
		\midrule
		SOI & Southern Oscillation Index. Atmospheric components of ENSO. Sea level pressure differences between Darwin and Tahiti, normalized \cite{Ropelewski1987} & monthly, since 1951 & CRU & NOAA PSL  \\
		\midrule
		ESPI & Precipitation anomalies in two areas: eastern tropical Pacific (10°S to 10°N, 160°E to 100°W) and Maritime Continent (10°S to 10°N, 90°E to 150°E), normalized \cite{Curtis2000}& monthly, since 1979  & GPCP & NOAA PSL \\
		\midrule
		200mb Zonal Winds & Zonal Winds equator anomalies (2.5°S to 2.5°N; 165°W to 110W°)  & monthly, since 1979& NCEP Reanalysis & NOAA PSL \\
		\bottomrule
	\end{tabularx}
	\caption{\label{tab:ensodata}Description of the time series data with corresponding sources. The data sources are either NOAA (National Oceanic and Atmospheric Administration) PSL (Physical Sciences Laboratory) or CPC (Climate Prediction Center).}
\end{table}

For our machine learning task, we consider four different data sets described in Table \ref{tab:ensodata}: sea-surface temperature anomolies (SST), southern oscillation index (SOI), ENSO Precipitation Index (ESPI) and zonal wind anomalies.  
Fig.~\ref{fig:timeseries}(c) shows time series of these data sets where time $t$ is in terms of months. 
The maxima of SST \cite{Trenberth1997} and ESPI \cite{Curtis2000} as well as the minima of SOI \cite{Rasmusson1983} and winds \cite{Bjerknes1969} (above or below a certain threshold) correspond to an El Ni\~no event.

The real-world ENSO system is clearly higher dimensional than the three-dimensional Lorenz and Rössler systems. 
While progress has been made in the use of RC for forecasting El Ni\~no events (for example, see \cite{jinno2025long,guardamagna2025explaining}), we do not focus on forecasting here. Instead, we consider the task of using only some observed data sets to predict another ``unseen'' observable of the ENSO system. For example, can the ESPI be accurately predicted using only the SST, SOI and wind data sets?

\section{Inference of unknown components}
\label{sec:results}

We focus on the inference of two unseen data components for the Lorenz and the Rössler chaotic systems and one component for ENSO. 
In each case we find the optimal performance of the NGRC algorithm, which depends on two hyperparameters: the number of time delayed data points fed into the network and the ridge regression parameter. Note that we fix the nonlinearity degree to $p=2$. We examine the influence of the hyperparameters on the performance of the NGRC as well as the influence of the discretization in time. For this purpose, the NGRC is optimized with respect to its performance in the testing phase. 

We compare the performance of the NGRC with traditional RC regarding its accuracy, the number of training points and the training time.
We use the same data sets for RC and NGRC. The data set is divided into a training and a testing part. The training data for RC and NGRC may differ since RC needs, in general, more training points for a good performance \cite{ngrc}. To measure the performance, the normalized root mean-squared error (NRMSE) of the resulting predicted time series and the normalized maximal distance is calculated via
\begin{align}
	\text{NRMSE} &\coloneq \dfrac{\sqrt{\dfrac{1}{n} \sum_{i=1}^n ||\hat{X_i} - X_{i}||^2}}{||X_\text{max} - X_\text{min}||},\\[.3cm]
	d_\text{max} &\coloneq  \frac{\max_{i=1,\ldots,n} ||\hat{x_i} - x_{i}||}{x_\text{max} - x_\text{min}}\label{eq:maxdistance}
\end{align}
where $\hat{X_i}$ is the (one- or two-dimensional) predicted state, $X_{i}$ is the observed state, $n$ is the length of the time series. The maximal distance is calculated for each component of the predicted state.

\begin{figure}[htp]
	\centering
	\includegraphics[width=\textwidth]{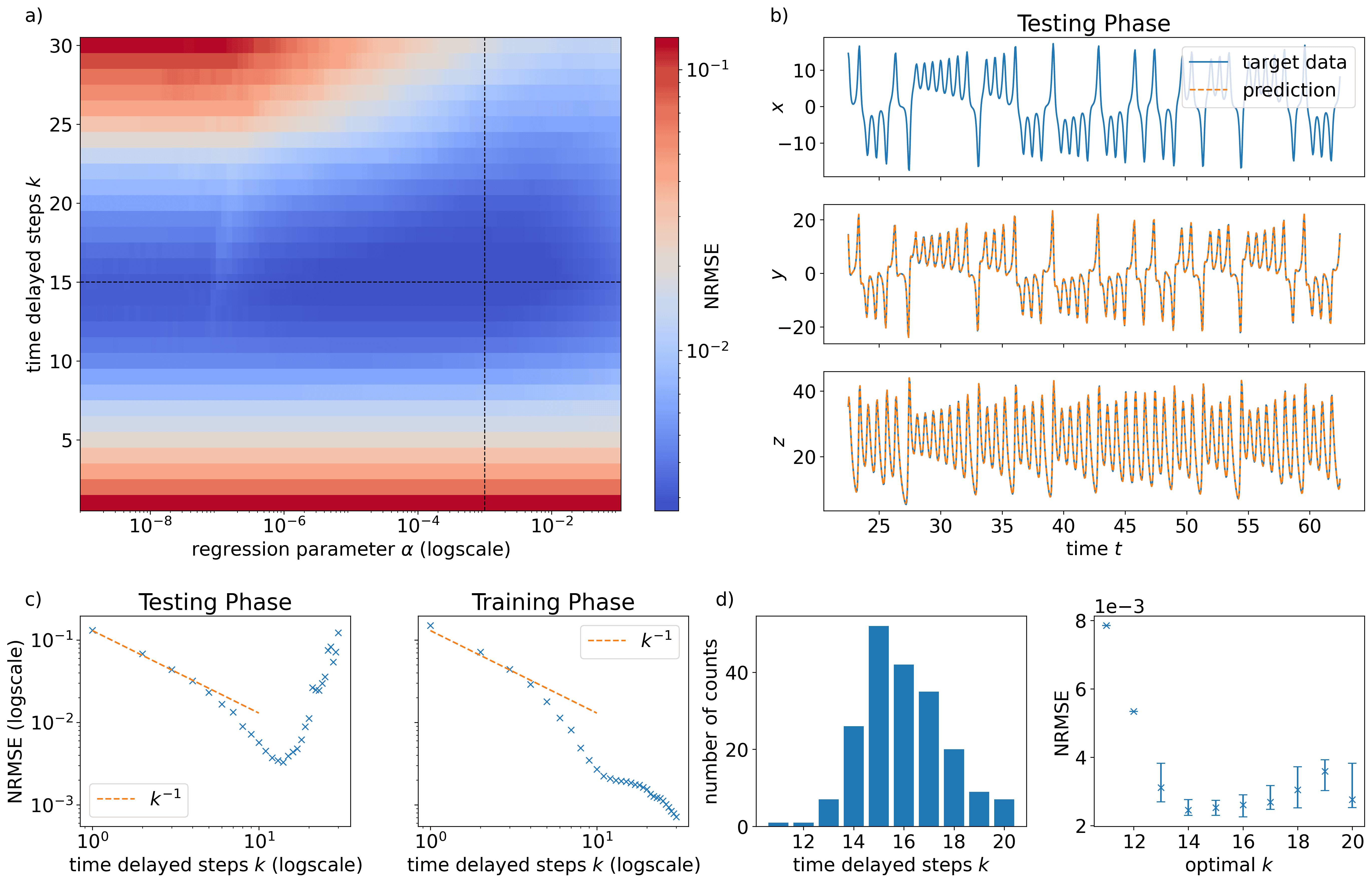}
	\caption{\label{fig:kvsalpha}NGRC applied to Lorenz system \eqref{eq:lorenz}. (a) The NRMSE of the testing phase plotted on a semi-logarithmic scale against parameters $\alpha \in [10^{-9}, 0.1]$ and $k \in [1,30]$. The plotted NRMSE is the median over 100 randomly chosen initial conditions for each parameter pair $(\alpha, k)$. The dashed lines indicate the optimal values for $\alpha$ and $k$. (b) Lorenz attractor with optimal NGRC performance for $\alpha =0.001$ and $k=15$, $\Delta t = 0.05$ and initial condition $(x_0, y_0, z_0) = (5.7,-1.5, 3.1)$. The blue curve shows the target data, the dashed orange curve the prediction. The corresponding testing error is $\text{NRMSE}\approx 3.92 \times 10^{-3}$. (c) NRMSE for the training and testing phases for different time delayed steps $k$ with fixed $\alpha$ = $0.001$ plotted on a logarithmic scale. The dashed line shows a decrease $\text{NRMSE}\sim k^{-1}$. Parameters and initial conditions: $\Delta t = 0.05$, $(x_0, y_0, z_0) = (-12,5, 22)$. (d) Distribution of the optimal $k$ values found for 200 random initial conditions, and the median of the NRMSE for each optimal $k$. The bars indicate the first and third quartile.}
\end{figure}

\subsection{Inference of two unknown components of the Lorenz system with NGRC}

\subsubsection{Performance and hyperparameter optimization}

In the following we analyze the performance of the NGRC at inferring the $y$ and $z$ components of the Lorenz system \eqref{eq:lorenz}, given its $x$ component.

First, we find the optimal hyperparameters $k$ (time delayed steps), $\alpha$ (ridge regression parameter) and $\Delta t$ (timestep). For this purpose, 50 warm-up points, 400 training points, and 800 testing points are selected successively from a given (numerically simulated) data set, in a similar fashion to \cite{ngrc}. 

A grid search is performed to find the optimal $k$ and $\alpha$, for which the NRMSE of the testing phase is minimal.
The NRMSE is calculated in Fig.~\ref{fig:kvsalpha}(a) for the $30\times 100$ grid of parameters $k \in [1,30]$, $\alpha \in [10^{-9}, 0.1]$ and fixed $\Delta t=0.05$. For each parameter pair, $k$ and $\alpha$, the NRMSE is calculated for 100 randomly chosen initial conditions ($(x_0,y_0) \in [-20,20]^2, z_0 \in [0, 50]$) with the plotted value representing the median. We observe a relatively large NRMSE for small values of $\alpha$, as well as for large and small values of $k$. 
The minimum is located at $k=15$ and  $\alpha =0.001$, as indicated by the two dashed lines in panel~(a). Panel~(b) demonstrates the inference of variables $y$ and $z$ based on $x$, using the initial condition $(x_0, y_0, z_0) = (-12,5,22)$. The inference does appear very successful, since the prediction of $y$ and $z$ in orange overlaps the target data in blue.

Figure~\ref{fig:kvsalpha}(c) shows the NRMSE for the training and testing phases as a function of the time delayed steps $k$ with fixed $\alpha=0.001$. For comparison, the dashed line shows a decrease $\text{NRMSE}\sim k^{-1}$. The minimum of the NRMSE for the testing phase corresponds to the optimal $k=15$. Overfitting can be observed for larger $k$, as the training error decreases while the testing error increases. 
To demonstrate the dependence on initial conditions, panel~(d) shows the distribution of optimal $k$ values found for 200 randomly chosen initial conditions  and a range of $k \in [1,20]$. This distribution peaks between 14 and 17. The optimal NRMSE lies in most of the cases near $3 \times 10^{-3}$ showing accurate prediction for different initial conditions.
Note that we take the median instead of the mean since the 200 initial conditions are chosen randomly and the algorithm fails in some special cases, where the training data is not representative of the chaotic attractor.

We now examine the influence of the timestep discretization, $\Delta t$, on the optimal value of time delayed steps $k$. For this, we consider 20 equally distributed values $\Delta t \in [0.01, 0.1]$, 30 values $k \in [1,30]$, and 50 different initial conditions for each $\Delta t$. 

\begin{figure}[!b]
	\centering
    \includegraphics[width=.8\textwidth]{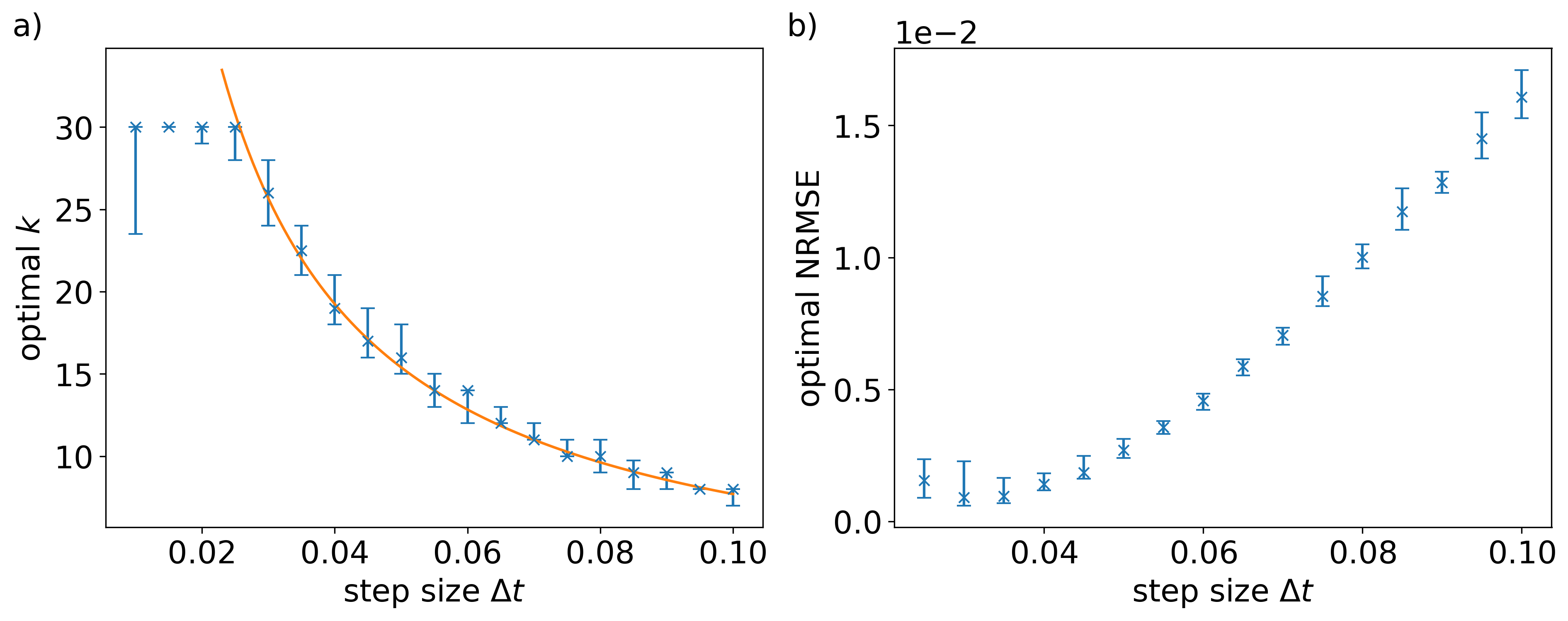}
	\caption{\label{fig:kvsdeltat} NGRC applied to Lorenz attractor. (a) The optimal number of time-delayed steps $k$ and (b) the corresponding NRMSE as a function of the discretization timestep $\Delta t$. For each $\Delta t$, the median and the first and third quartiles are calculated over 50 randomly fixed initial conditions each time step. The orange curve shows ${k \Delta t} = 0.77$.}
\end{figure}

Figure~\ref{fig:kvsdeltat}(a) displays the optimal $k$ for each $\Delta t$, while panel~(b) shows the corresponding NRMSE in terms of the median and the first and third quartile. The horizontal axis is cut off for small values of $\Delta t$ in panel~(b), since we set an upper limit of $k=30$, resulting in large MSE values in panel~(b) and the plateau in panel~(a) for small time steps. In panel~(a) we observe a clear trend for $\Delta t>0.02$, indicating how the optimal time-delayed steps is inversely proportional to the timestep size. By assuming an inverse proportional dependence, we obtain the mean value with standard deviation of $k\Delta t$ as 
\begin{align}
	\overline{k \Delta t} = 0.77 \pm 0.06,
\end{align}
plotted in orange in Fig.~\ref{fig:kvsdeltat}~(a). This value represents the amount of memory that the NGRC needs in order to correctly infer the unknown components.

\subsubsection{Comparison of NGRC and RC}

First, we describe the training and testing setups for NGRC and RC. The same 800 data points were used for testing the RC and NGRC, however the training sets are different. Since hyperparameter optimization for RC has already been performed by \cite{rc_lu2}, where the $y$ and $z$ components were predicted for a given $x$ input, we use the hyperparameters for RC from \cite{rc_lu2}. For the ridge regression parameter, an optimal value of $10^{-9}$ was estimated by considering 5 different initial conditions for 15 regression values and 30 trials each. The RC training was conducted using 5200 \cite{rc_lu2} data points to facilitate accurate training, and also using 400 training points for a more direct comparison with the NGRC.

Table \ref{tab:lorenzsum} summarizes the results. The median and the 95\% confidence interval are calculated over 100 initial conditions. The computation time is the training and prediction time using data from a single initial condition 100 times, then divided by 100. This is done 7 times to calculate a mean and standard deviation. Generally, the RC with 5200 training points performs slightly better than the NGRC regarding both the NRMSE and the maximal distance. It should be noted that the training errors are less comparable since the training sets are different. However, the NGRC requires only 1/13 of the training points compared to the RC algorithm resulting in much larger computational times for the RC, even considering the same training set of 400 points.

\begin{table}[!htp]
	\centering
	\begin{tabular}{ c|c|c|c|c|c||c } 
		  & &
         \makecell{$\text{NRMSE}_\text{test}$\\$\times 10^{-4}$} &
         \makecell{$\text{NRMSE}_\text{train}$\\$\times 10^{-4}$} &
         \makecell{$y_\text{max}$\\$\times 10^{-3}$} &
         \makecell{$z_\text{max}$\\$\times 10^{-3}$} &
         \makecell{comp.\\time}\\
         \hline \hline
         
		\multirow{2}{*}{\makecell{NGRC\\ $[400]$}} & median & $27$ & $17$& $22$ & $ 18 $ & \multirow{2}{*}{\makecell{$17$\\ $\pm 0.1$ms}}\\
        & 95\% & $[21, 99] $ & $[10, 22] $& $[12, 78] $ & $[4, 62]$ & \\ \hline

		\multirow{2}{*}{\makecell{RC\\ $[5200]$}}& median &$2.8$ & $2.2$& $0.2 $ & $4.0$& \multirow{2}{*}{\makecell{$635$\\ $\pm 17$ms}}\\ 
        & 95\% & $[1.2, 17]$ & $[1.0, 5.2]$&  $[0.06, 0.5]$ & $[1.2, 37]$\\\hline
        
		\multirow{2}{*}{\makecell{RC\\ $[400]$}} & median & $13.5$ & $0.5$ & $1.1$  & $26$& \multirow{2}{*}{\makecell{284\\ $\pm 11$ms}}\\ 
        & 95\% & $[3.8, 166]$ & $[0.2, 3.5] $& $[0.2, 6.5]$ & $[6.4, 186]$
	\end{tabular}
	\caption{\label{tab:lorenzsum}Comparison of NRMSE, normalized maximal distance (\ref{eq:maxdistance}) and computation time for the Lorenz system using the NGRC with 400 training points and RC with 400 and 5200 training points. The number of testing points is 800 in all cases. The medians and the 95\% confidence intervals are calculated over 100 different initial conditions. The average computation time is calculated for the initial condition $(x_0, y_0, z_0) = (-12,5, 22)$. Timestep is $\Delta t = 0.05$.}
\end{table}

\subsection{Inference of two unknown components of the Rössler system with NGRC}

\subsubsection{Performance and hyperparameter optimization}
We apply a similar framework for the analysis of the NGRC performance, with respect to the parameters $k$, $\alpha$ and $\Delta t$, for the Rössler system \eqref{eq:roessler} as we did for the Lorenz system.
We choose 50 warm-up points, 600 training points and 1000 testing points successively from a given dataset. We consider a given $x$ component and use the trained NGRC to infer $y$ and $z$ components.
 
\begin{figure}[!b]
	\centering
	\includegraphics[width=\textwidth]{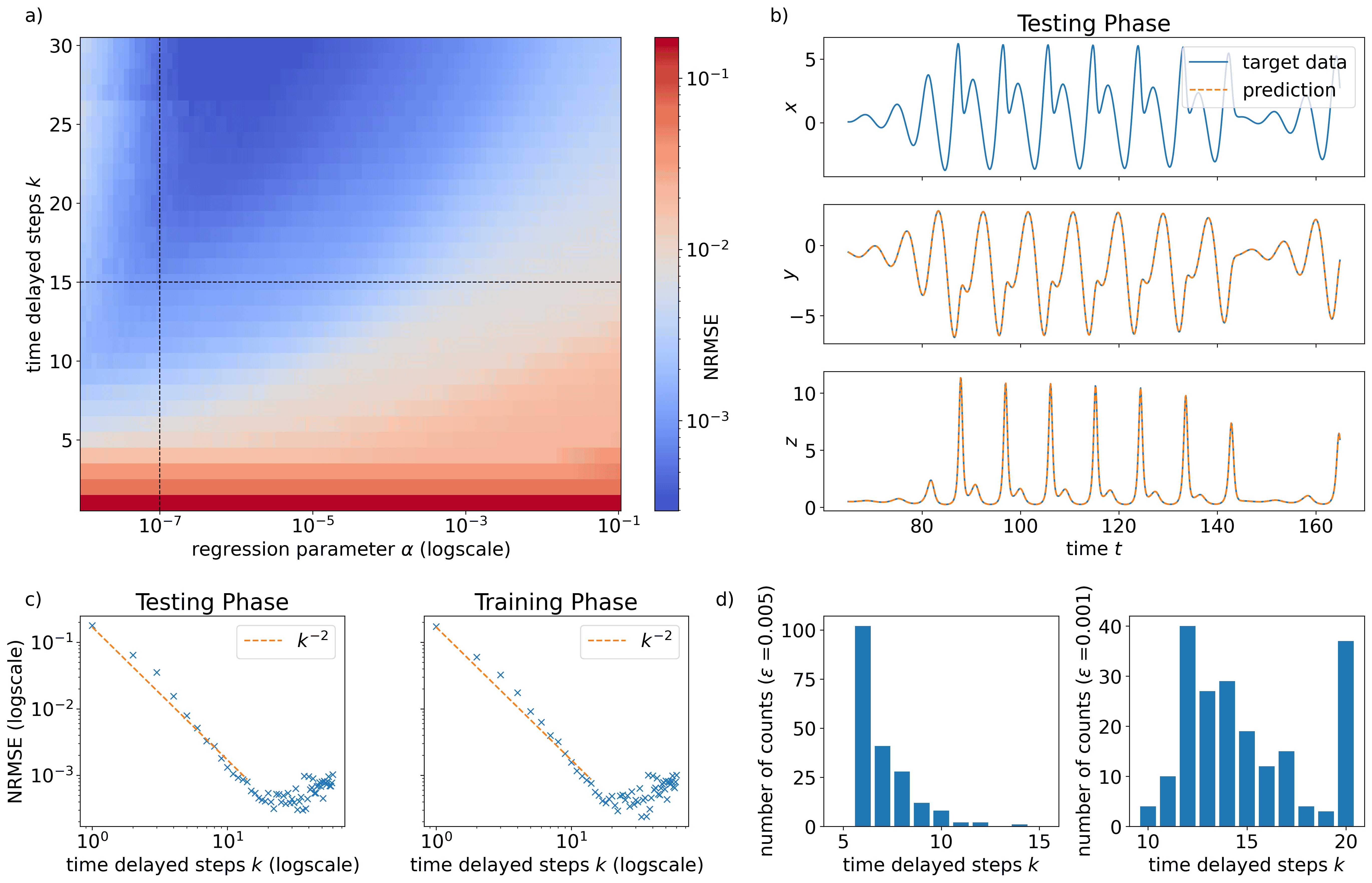}
	\caption{\label{fig:3D_roessler}(a) NGRC applied to Rössler system (\ref{eq:roessler}) with stopping threshold $\varepsilon = 0.001$. The NRMSE of the testing phase is plotted on a semi-logarithmic scale against the parameters $\alpha \in [10^{-8}, 0.1]$ and $k \in [1,30]$. The plotted NRMSE is the median over 100 randomly chosen initial conditions for each parameter pair $(\alpha,k)$. (b) Inference of $y$ and $z$ with optimal NGRC performance for $\alpha = 10^{-7}$ and $k=15$, $\Delta t = 0.1$, and initial condition $(x_0, y_0, z_0 ) = (-0.6,  0.2, -0.1)$. The target data is shown in blue, the prediction with dashed orange. The testing error is $\text{NRMSE} \approx 5.9 \times 10^{-4} $. (c) NRMSE for the training and testing phase plotted on a logarithmic scale versus the time delayed steps $k\in [1,60]$ with $\alpha = 10^{-7}$ fixed. The dashed line shows $\text{NRMSE} \sim k^{-2}$. Parameters and initial condition: $(x_0, y_0, z_0 ) = (-0.6 , 0.2 ,-0.1)$, $\Delta t = 0.1$. (d) Distribution of the optimal $k$ values found for 200 random initial conditions for two different stopping thresholds $\varepsilon = 0.005$ and $\varepsilon = 0.001$, $\alpha = 10^{-7}$.}
\end{figure}

A grid search is performed to find the optimal $\alpha$ and $k$, where the time step size is fixed as $\Delta t = 0.1$. The NRMSE is calculated in Fig.~\ref{fig:3D_roessler}(a) for a $100\times30$ grid of parameters $\alpha \in [10^{-8}, 0.1]$ and $k\in [1,30]$. For each pair of parameters, the median of the NRMSE is shown for 100 randomly chosen initial conditions ($(x_0, y_0, z_0) \in [-1,1]^3$). In contrast to the Lorenz system, the NRMSE decreases for further increasing $k$ without an overfitting effect, at least until $k=30$. In other words, there is no interior minimum of NRMSE in Fig.~\ref{fig:3D_roessler}(a), as there is in Fig.~\ref{fig:kvsalpha}(a), instead there is only a boundary minimum at $k=30$. Hence, instead of selecting $30$ as the optimal $k$ value, we select the ($\alpha$, $k$) pair that satisfy a certain threshold
\begin{align*}
    \text{NRMSE}_\text{test} \leq \varepsilon.
\end{align*}
while minimising $k$. We call this value $\varepsilon$ the \textit{stopping threshold}. For Fig.~\ref{fig:3D_roessler}, we choose a stopping threshold of $\varepsilon = 0.001$ which is satisfactory for the NRMSE in our setting. Under this assumption, the optimal hyperparameter setting is $\alpha = 10^{-7}$ and $k=15$, as indicated by the dashed lines, which provides the minimal value of $k$ with a NRMSE equal to or less than 0.001. Panel~(b) illustrates the performance of the NGRC with the optimal hyperparameter setting and the initial condition $(x_0, y_0, z_0 ) = (-0.6,  0.2, -0.1)$. Again, the prediction of the inferred variables in orange overlap with the target data, indicating that 0.001 is a sufficient stopping threshold.

Fig.~\ref{fig:3D_roessler}(c) shows the influence of varying the number of time delayed steps, $k\in[1,60]$, on the NRMSE.
For comparison, the dashed line represents $\text{NRMSE}\sim k^{-2}$. 
Additionally, panel~(d) shows the distribution of optimal $k$ values for $\alpha = 10^{-7}$ and two different stopping thresholds $\varepsilon = 0.005$ and $\varepsilon = 0.001$. An optimal $k$ is found for 200 randomly chosen initial conditions. The optimal $k$ peaks at $k= 6$ for the larger threshold and lies mostly between 12 and 15 for the smaller one. This suggests that, since the size of the feature vector scales quadratically with $k$, the computational effort can be significantly reduced by slightly lowering the stopping threshold.

\begin{figure}[htp]
	\centering
		\includegraphics[width=.8\textwidth]{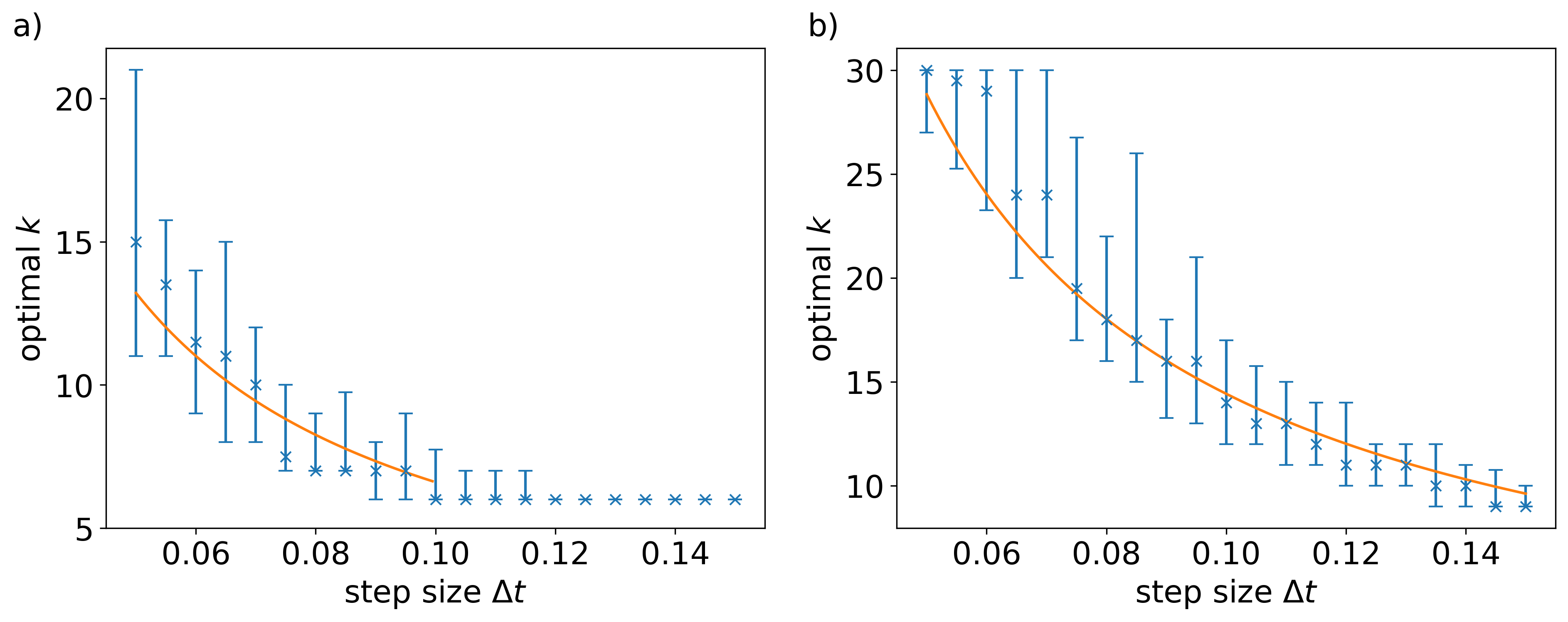}
	\caption{\label{fig:kvsdeltat_roessler} NGRC applied to Rössler attractor depicting the influence of the step size $\Delta t$ on the optimal $k$. (a) Stopping threshold (a) $\varepsilon = 0.005$ and (b) $\varepsilon = 0.001$. For each $\Delta t$, the median is calculated over 50 randomly chosen initial conditions, and the bars indicate the first and third quartile. The orange curve shows ${k\Delta t} = 0.66$ for $\Delta t \in [0.05, 0.1]$ in (a) and ${k\Delta t} = 1.44$ in (b).}
\end{figure}

To examine the influence of $\Delta t$ on the optimal $k$, we take 20 equidistantly distributed values for $\Delta t \in [0.05, 0.15]$, 30 values for $k\in [1,30]$ and 50 initial conditions for each time step. Fig.~\ref{fig:kvsdeltat_roessler} shows the results for two different stopping thresholds, $\varepsilon =0.005$ in panel~(a) and $0.001$ in panel~(b). We find the inverse proportional dependence 
\begin{align*}
	\overline{k\Delta t} = 1.44 \pm 0.11
\end{align*}
for the stopping threshold $\varepsilon =0.001$. Increasing the stopping threshold to $\varepsilon = 0.005$ reduces the value to $\overline{k\Delta t} = 0.66 \pm 0.10$ for $\Delta t \in [0.05, 0.1]$. Hence, fewer delayed time steps $k$ are sufficient when a lower prediction accuracy is tolerated.

\subsubsection{Comparison of NGRC and RC}

Analogously to the Lorenz system, the data points are chosen such that the testing phases of both algorithms take the same points to enable a comparison between the two algorithms. However, the training sets may differ. We take the parameter values from \cite{rc_lu2} and optimize the RC with respect to the regression parameter by choosing the same stopping threshold and considering 5 different initial conditions with 30 trails each. The training was conducted using 2600 training points, and a further comparison to NGRC was made by using 600 training points.

Table \ref{tab:roesslersum} summarizes the results. The NRMSE for both algorithms (NGRC with 600 and RC with 2600 training points) lie in the same range, since both are implemented to stop at this threshold. However, the NGRC algorithm shows larger deviations in the 90\% confidence interval. Nevertheless, to achieve this accurate prediction, a much smaller training data set is necessary for the NGRC algorithm. As for the Lorenz attractor, the RC is performed on the same training data set as the NGRC resulting in a much larger NRMSE in the testing phase. In this setting, the testing NRMSE does not achieve the desired threshold of $1 \times 10^{-3}$. The normalized maximal distance highlights the better performance of the NGRC algorithm. For both components, the NGRC has the smallest normalized maximal distance compared to the RC method with 2600 and 600 training points.  

Moreover, Table \ref{tab:roesslersum} highlights the average computation time needed to calculate the prediction. The computation time is much smaller for the NGRC as for the RC algorithm, even if the training data set is the same. 

\begin{table}[!ht]
	\centering
    \begin{tabular}{ c|c|c|c|c|c||c } 
		  & &
         \makecell{$\text{NRMSE}_\text{test}$\\$\times 10^{-4}$} &
         \makecell{$\text{NRMSE}_\text{train}$\\$\times 10^{-4}$} &
         \makecell{$y_\text{max}$\\$\times 10^{-3}$} &
         \makecell{$z_\text{max}$\\$\times 10^{-3}$} &
         \makecell{comp.\\time}\\
         \hline \hline
         
		\multirow{2}{*}{\makecell{NGRC\\ $[600]$}} & median & $9.0$ & $5.1$& $6.8$ & $ 5.4 $ & \multirow{2}{*}{\makecell{$21$\\ $\pm 0.4$ms}}\\
        & 95\% & $[5.7, 80] $ & $[3.2, 6.1] $& $[2.6, 82] $ & $[2.0, 67]$ & \\ \hline

		\multirow{2}{*}{\makecell{RC\\ $[5200]$}}& median &$10 $ & $8.9$& $8.6 $ & $6.7$& \multirow{2}{*}{\makecell{$491$\\ $\pm 85$ms}}\\ 
        & 95\% & $[4.4, 15]$ & $[5.6, 13]$&  $[2.5, 14]$ & $[1.9, 10]$\\\hline
        
		\multirow{2}{*}{\makecell{RC\\ $[600]$}} & median & $17$ & $10$ & $14$  & $5.4$& \multirow{2}{*}{\makecell{294\\ $\pm 0.8$ms}}\\ 
        & 95\% & $[6.9, 40]$ & $[5.4, 18] $& $[4.0, 29]$ & $[3.1, 22]$
	\end{tabular}
	\caption{\label{tab:roesslersum}Comparison of NRMSE, normalized maximal distance \ref{eq:maxdistance} and computation time for the R\"ossler system using the NGRC with 600 training points and RC with 600 and 2500 training points. The number of testing points is in all cases 1000. The median and the 95\% confidence interval are calculated over 100 different initial conditions. The time average is calculated with the standard deviation over 7 runs, 100 loops each. Timestep $\Delta t = 0.1$ and initial condition $(x_0, y_0, z_0 ) = (-0.6,  0.2, -0.1)$.}
\end{table}

\subsection{Inference of one Component of the El Ni\~no--Southern Oscillation}

In this section we consider the ENSO as a climate application with real-world data sets. Given that the data points are observed on a monthly basis, there are only 516 available data points for the shortest time series which dates back to 1979. We take the same number of warm-up, training and testing points for both algorithms: 60 warm-up points, 180 training points and 276 testing points. To simplify the algorithms, the reservoir parameters for the RC algorithm are kept the same as for the Lorenz system. Since there is no visible improvement for higher $k$, only values of $k$ between 1 and 10 are chosen. Additionally, 50 values of $\alpha\in [0.1, 10^{-8}]$ are considered. 

\begin{table}[!ht]
	\centering
	\begin{tabularx}{\textwidth}{ P{1.3cm}|P{1.2cm}|P{1.7cm}|P{2cm}|P{2cm}|P{.9cm}|P{1.9cm}|P{1cm} } 
		input data & target data & algorithm & $\text{NRMSE}_\text{test}$ $\times 10^{-1}$ & $\text{NRMSE}_\text{train}$ $\times 10^{-1}$& opt. $k$ & time & $d_\text{max}$ $\times 10^{-1}$\\ \hline \hline
		
		\multirow{3}{*}{\makecell{SOI,\\ SST, \\ winds}} & \multirow{3}{*}{ESPI} & NGRC & $0.9  $ & $1.0  $ & 1& 383 $\mu$s\hspace{.5cm} $\pm$ 9 $\mu$s & 3.6\\ \cline{3-8}
		& &RC & $1.1 $ & $1.0 $ & -- & 240 ms\hspace{0.5cm} $\pm$ 2 ms & 4.7 \\ \cline{1-8}
		
		\multirow{3}{*}{\makecell{SST,\\ ESPI,\\ winds}}
        & \multirow{3}{*}{SOI} & NGRC & $1.1 $ & $1.2 $ & 1& 389 $\mu$s \hspace{0.5cm}$\pm$ 3 $\mu$s & 4.1 \\ \cline{3-8}
		& &RC & $1.3 $ & $1.2 $ & -- & 241 ms\hspace{0.5cm} $\pm$ 0.5 ms & 4.1 \\ \cline{1-8}
		
		\multirow{3}{*}{\makecell{SOI,\\ ESPI,\\ winds}} & \multirow{3}{*}{SST} & NGRC & $1.1 $ & $0.8 $ & 2& 1.1 ms \hspace{0.5cm}$\pm$ 11 $\mu$s & 3.5 \\ \cline{3-8}
		& &RC & $1.0 $ & $0.8$ & --& 242 ms\hspace{0.5cm} $\pm$ 0.6 ms & 3.4 \\ \cline{1-8}

        \multirow{3}{*}{\makecell{SOI,\\ ESPI,\\ SST}} & \multirow{3}{*}{winds} & NGRC & $1.5 $ & $1.2 $ & 1& 390 $\mu$s \hspace{0.5cm} $\pm$ 6 $\mu$s & 4.4 \\ \cline{3-8}
		& &RC & $1.4 $ & $1.1$ & --& 240 ms\hspace{0.5cm} $\pm$ 0.5 ms & 5.4 \\
		
	\end{tabularx}
	\caption{\label{tab:ensosum}Comparison of NRMSE, normalized maximal distance \ref{eq:maxdistance} and computational time for the inference of different ENSO observables. We take 180 training points and 276 testing points. The optimal value for $\alpha$ is $0.1$ in all cases. The average computation time with standard deviation is calculated over 7 runs, each 100 loops.}
\end{table} 

\begin{figure}[!t]
	\centering
	\begin{subfigure}{\textwidth}
		\centering
		\includegraphics[width=.8\textwidth]{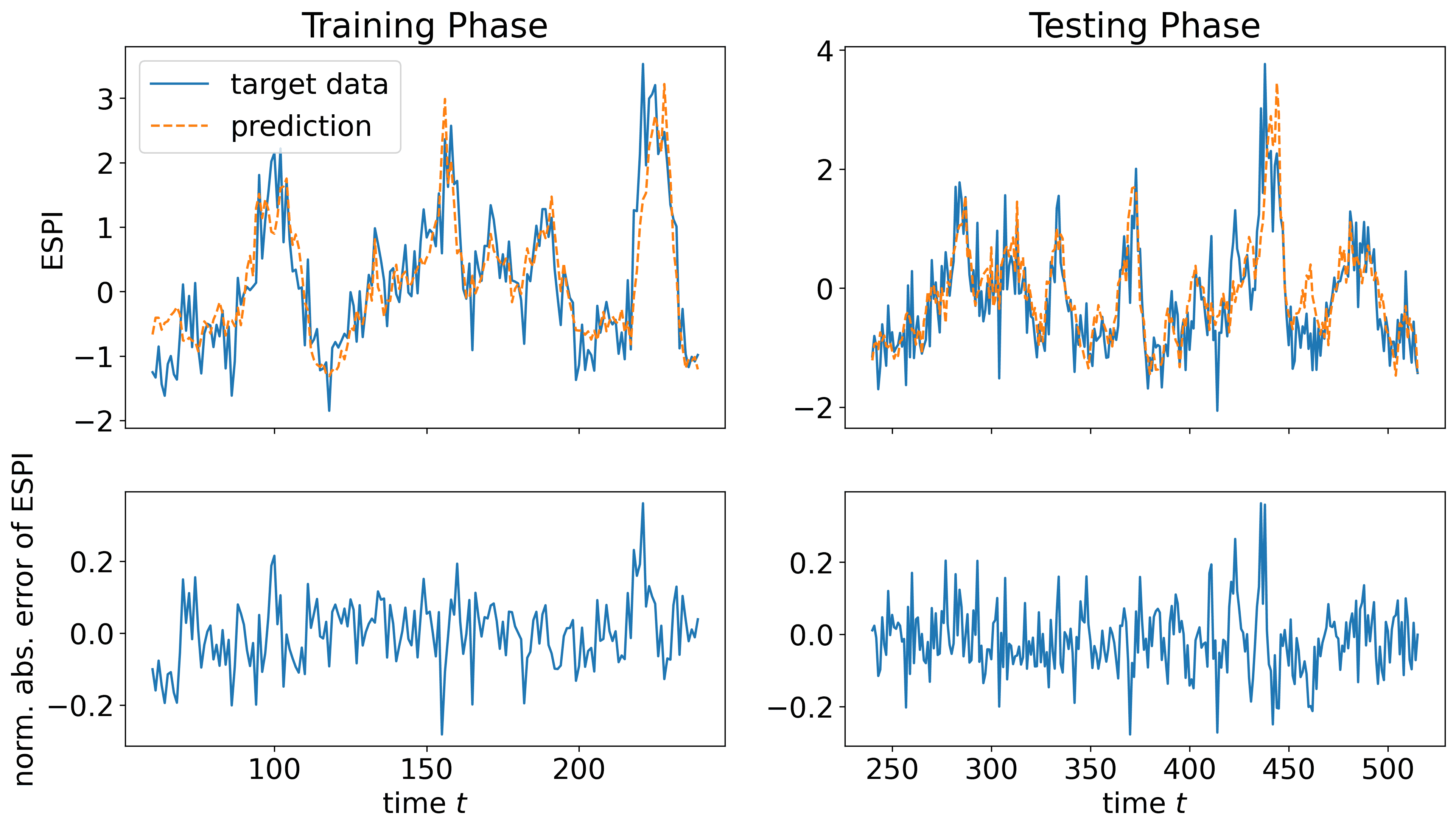}
		\caption{Optimal NGRC performance, $\text{NRMSE}_\text{NGRC} = 9.6 \times 10^{-2}$. Top panel: The target is shown in blue, the prediction with dashed orange. Bottom panel: normalized absolute error in each time step.}
	\end{subfigure}
	\vspace{0.5cm}
	\begin{subfigure}{\textwidth}
		\centering
		\includegraphics[width=.8\textwidth]{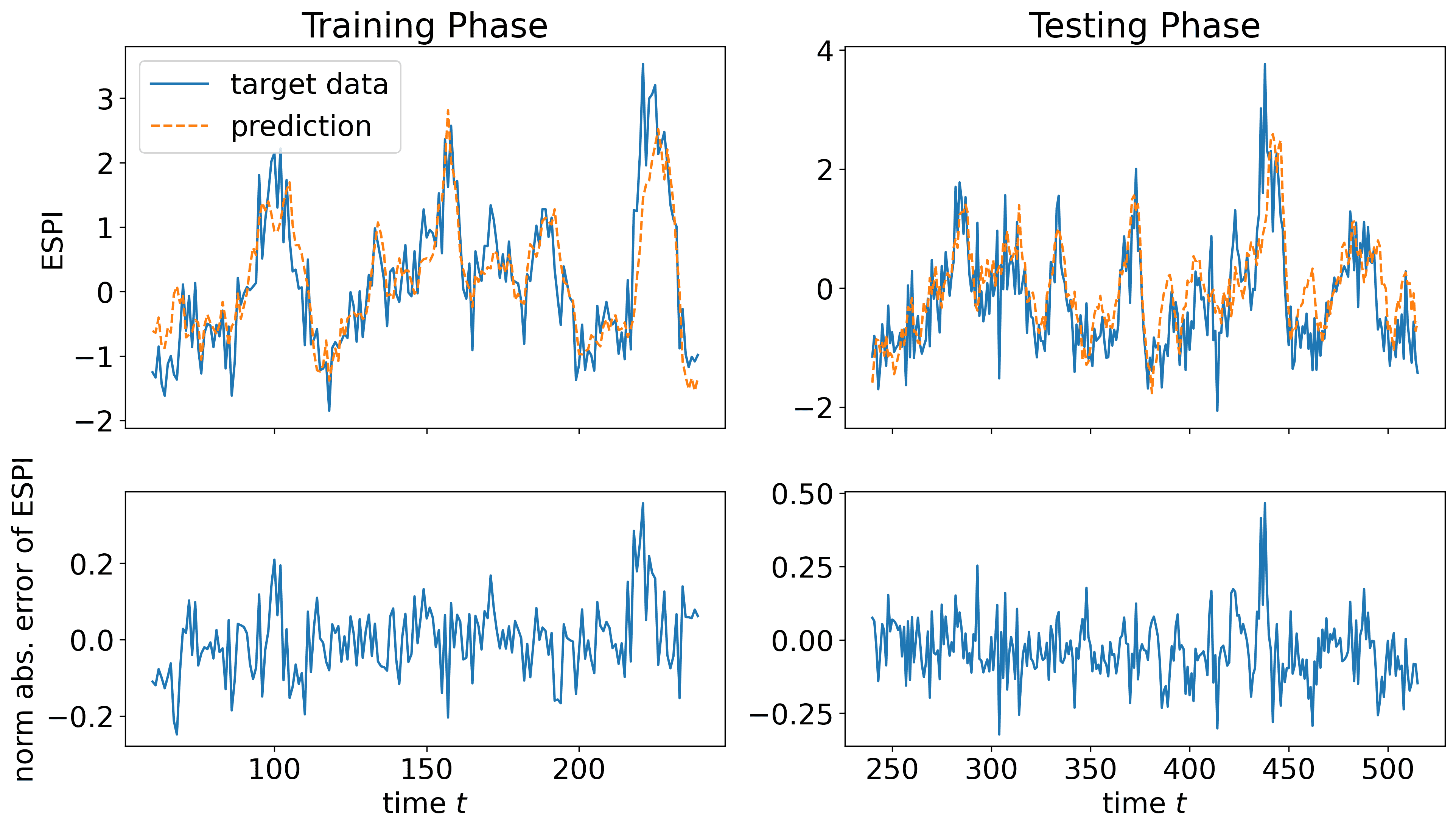}
		\caption{Optimal RC performance, $\text{NRMSE}_\text{RC} = 2.1 \times 10^{-1}$. Top panel: The target is shown in blue, the prediction with dashed orange. Bottom panel: normalized absolute error in each time step.}
	\end{subfigure}
	\caption{\label{fig:optplot_enso_espi}ENSO system inferring ESPI. $\Delta t = 1$ represents monthly time steps. The left side shows the training, the right side the testing phase which are the same for both algorithms.}
\end{figure}

Table \ref{tab:ensosum} summarizes the different data sets used for training and testing, along with the resulting NRMSE values. Although the results are less accurate than for the Lorenz and Rössler systems, the model still shows a promising performance, especially given the complexity of real-world ENSO data. Fig.~\ref{fig:optplot_enso_espi} and \ref{fig:optplot_enso_soi}--\ref{fig:optplot_enso_winds} illustrate the inference of the different observables with the predictions in orange and the true target data in blue. In all cases, the predictions reflect key trends of the underlying dynamics quite well. Smaller-scale features are not yet fully reproduced. This is reflected in the fluctuations of the normalized absolute error, which is calculated for each time step as
\begin{align}
    e_\text{abs} \coloneq \frac{\hat{x}-x}{x_\text{max}-x_\text{min}},
\end{align}
where $x$ is the observed data series and $\hat{x}$ is the predicted time series. This error is usually between $\pm 0.2$ and indicates room for improvement.

Comparing the algorithms, we note that the NGRC algorithm performs similar to the RC algorithm in the testing phase for all experiments. Again, the time needed to calculate the results is consistently and significantly smaller for the NGRC than for the RC algorithm.

\section{Discussion}

In this study we have demonstrated the accurate inference of two unknown components of the Lorenz and Rössler systems using the NGRC algorithm. The NRMSE is very small and there are no visible differences between the inferred values and the target data. Compared to the traditional RC algorithm, the NGRC algorithm required less training data and computation time. Given the simpler architecture of the NGRC algorithm, a shorter training time was expected. 

We considered different data sets when calculating the optimal number of time delayed steps $k$, sampled for 100 different initial conditions. For the Lorenz system, the optimal number of time delayed steps $k$ was in most cases between 14 and 17, and for the Rössler system between 12 and 15 for a stopping threshold $\varepsilon = 0.001$. This showed that the NGRC algorithm shows an accurate performance also for different initial conditions. However, the values for the number of time delayed steps was unexpected as \cite{ngrc} only considered $k \leq 4$. Our findings revealed a decrease of order $-1$ (Lorenz) and of order $-2$ (Rössler) in the NRMSE as $k$ increases, emphasizing the importance of considering a larger number of time delayed steps for more accurate predictions. Interestingly, these trends begin to falter for very large $k$ where the NRMSE actually begins to increase.

We also showed the expected inversely proportional dependence between $k$ and $\Delta t$, which demonstrates that the physical time span covered by the time delayed steps, rather than the number of steps itself, is crucial for the prediction accuracy of the NGRC model. For the Lorenz system, we find $k\Delta t = 0.76 \pm 0.05$. With coarser time step sizes, fewer time delayed steps are required since a larger segment of the curve is covered. However, to achieve a lower NRMSE, finer discretization is necessary for a more detailed representation of the dynamics, leading to an increase in the number of required time delayed steps to adequately capture the system's dynamical behavior. However, a large number of time delayed steps results in higher computational costs, since the dimension of the feature vector increases significantly. Therefore, we found that selecting an appropriate time step size with a corresponding value of $k$ is crucial, based on the desired task's objectives. For the Rössler system with a stopping threshold $\varepsilon = 10^{-4}$ we find $k\Delta t = 1.44 \pm 0.11$. With coarser time step sizes, fewer time delayed steps are required, since a larger time span of the dynamics is captured. Even for very coarse time step sizes, an NRMSE below $10^{-4}$ is still reached. Therefore, to lower the computational costs, coarser time step sizes and therefore smaller $k$ needs to be chosen.

Overall, neither algorithm consistently outperforms the other considering all metrics (NRMSE, maximal distance and computational time). The performance of NGRC and RC depends on the specific dynamical system and the evaluation metric. While RC achieves slightly lower prediction errors for the Lorenz system, the NGRC reaches slightly better prediction errors for the R\"ossler system. 
However, the simpler NGRC algorithm needs less training points and computational time for the results in all experiments.   

For the ENSO system, we achieved promising results, with the NGRC performing slightly better than the traditional RC. The same level of accuracy as the other two dynamical systems could not be achieved. Arguably, this was to be expected as the ENSO system is a real-world application and not a system described by a known ODE. Therefore, the underlying dynamics are mathematically not as well understood as for the other two and the measured data sets include noise, which generally requires tailored solutions to ensure accurate forecasting and inference (e.g. \cite{gottwald2021combining}).

In conclusion, we have shown that the NGRC is a powerful tool for the inference of unknown parameters in chaotic systems. The results indicate that the choice between NGRC and RC involves a trade-off between accuracy and computational efficiency. As complex and chaotic behavior is a common feature in many dynamical systems, the NGRC algorithm has the potential to be an effective, accurate and fast solution for inferring observables across a wide range of applications.

\section*{Acknowledgments}
This work emanated from the research funded by Taighde Éireann-Research Ireland (Grant No. FFPA/\allowbreak12066). We acknowledge Florian Stelzer for helpful discussions and for providing his reservoir computing code. The NGRC code used in this work is based on the implementation provided by \cite{ngrc}.

\bibliographystyle{elsarticle-num}
\bibliography{bibliography}
\biboptions{sort&compress}

\newpage
\appendix

\section{Inferring components of ENSO}

\begin{figure}[!h]
	\centering
	\begin{subfigure}{\textwidth}
		\centering
		\includegraphics[width=.8\textwidth]{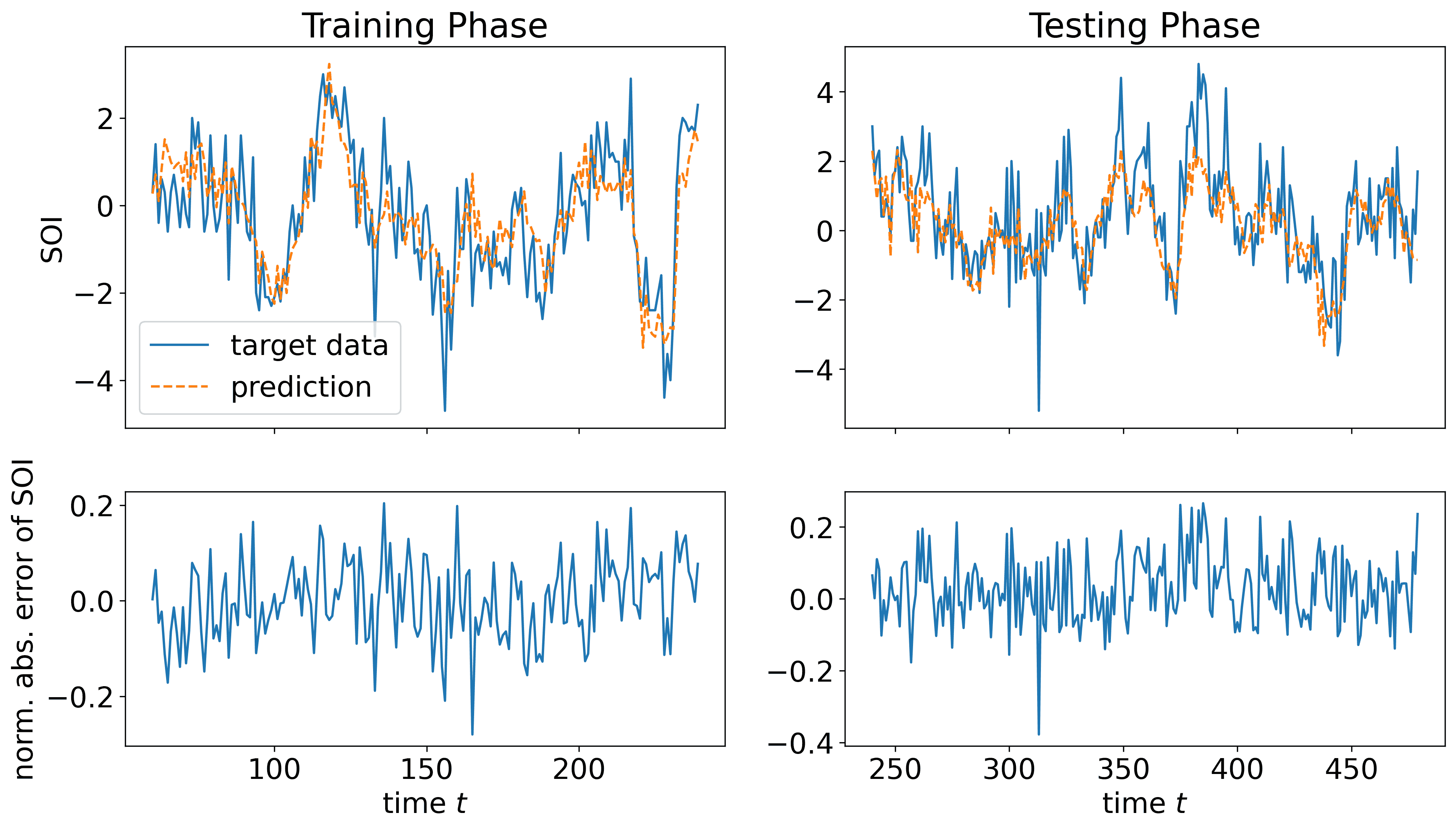}
		\caption{Optimal NGRC performance, $\text{NRMSE}_\text{NGRC} = 1.1 \times 10^{-1}$. Top panel: The target is shown in blue, the prediction with dashed orange. Bottom panel: normalized absolute error in each time step.}
	\end{subfigure}
	\vspace{0.5cm}
	\begin{subfigure}{\textwidth}
		\centering
	       \includegraphics[width=.8\textwidth]{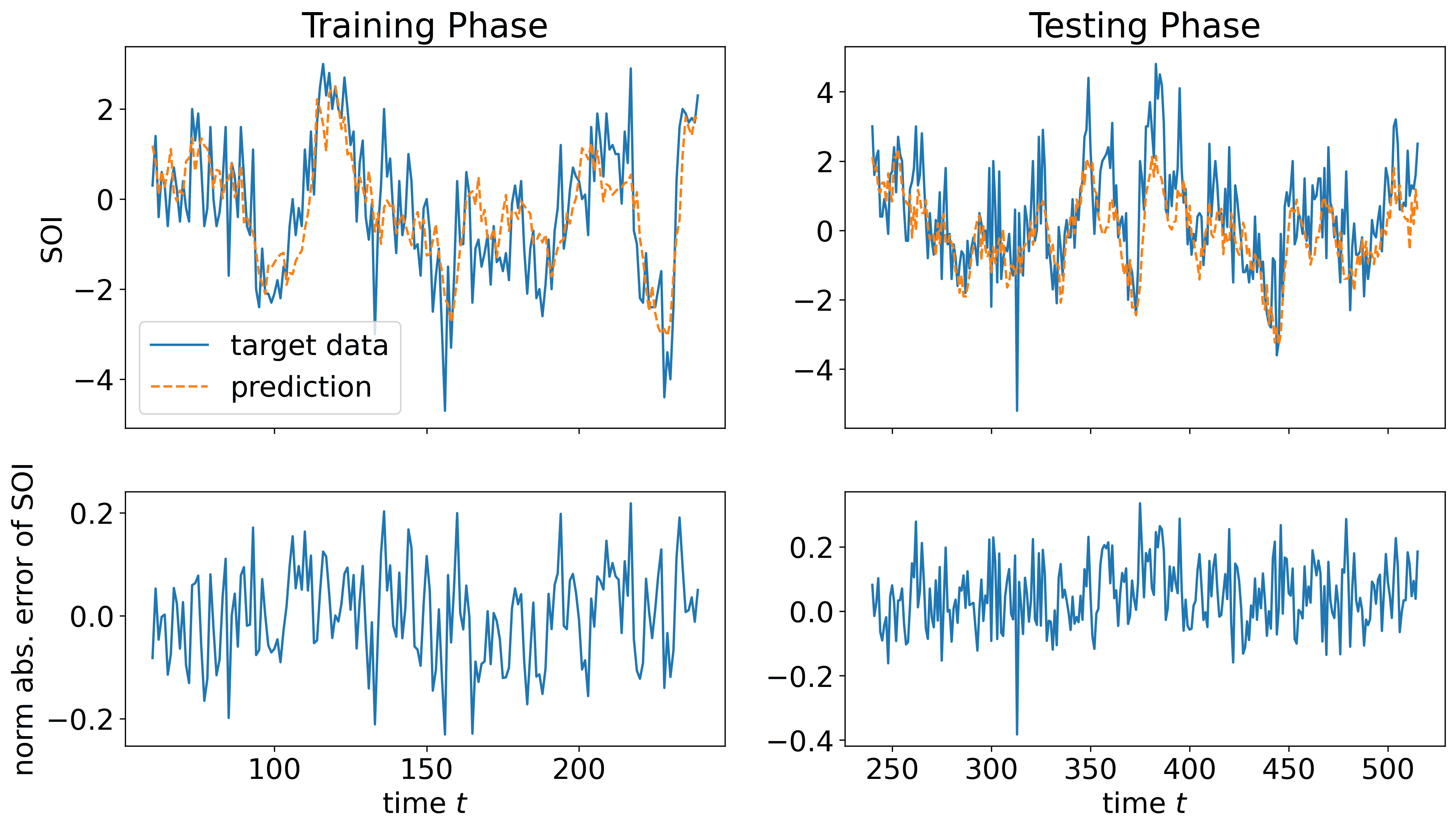}
		\caption{Optimal RC performance, $\text{NRMSE}_\text{RC} = 2.0 \times 10^{-1}$. Top panel: The target is shown in blue, the prediction with dashed orange. Bottom panel: normalized absolute error in each time step.}
	\end{subfigure}
	\caption{\label{fig:optplot_enso_soi}ENSO system inferring SOI. $\Delta t = 1$ represents monthly time steps. The left side shows the training, the right side the testing phase which are the same for both algorithms.}
\end{figure}
 
\begin{figure}[!htp]
	\centering
	\begin{subfigure}{\textwidth}
		\centering
		\includegraphics[width=.8\textwidth]{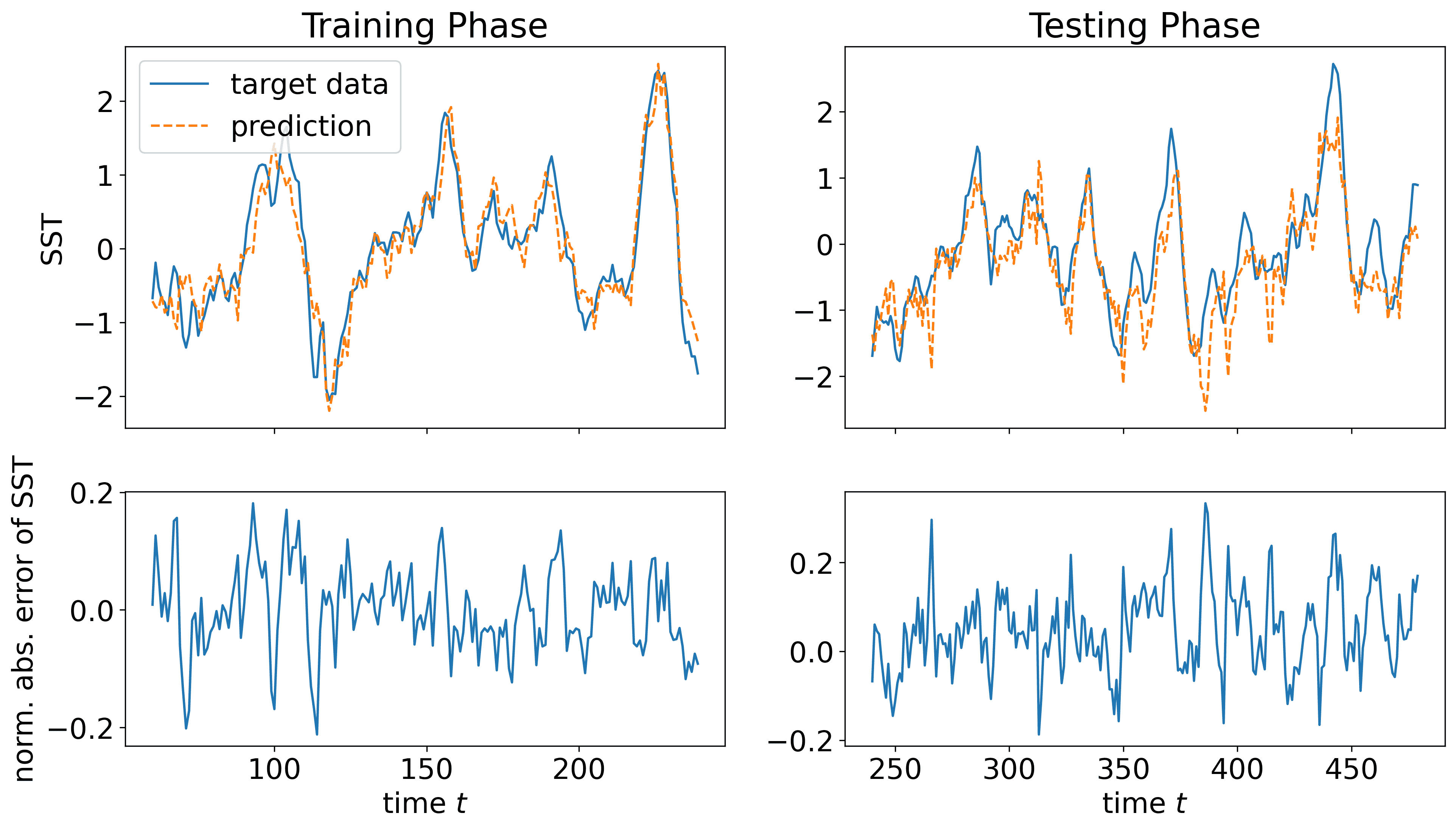}
		\caption{Optimal NGRC performance, $\text{NRMSE}_\text{NGRC} = 1.1 \times 10^{-1}$. Top panel: The target is shown in blue, the prediction with dashed orange. Bottom panel: normalized absolute error in each time step.}
	\end{subfigure}
	\vspace{0.5cm}
	\begin{subfigure}{\textwidth}
		\centering
		\includegraphics[width=.8\textwidth]{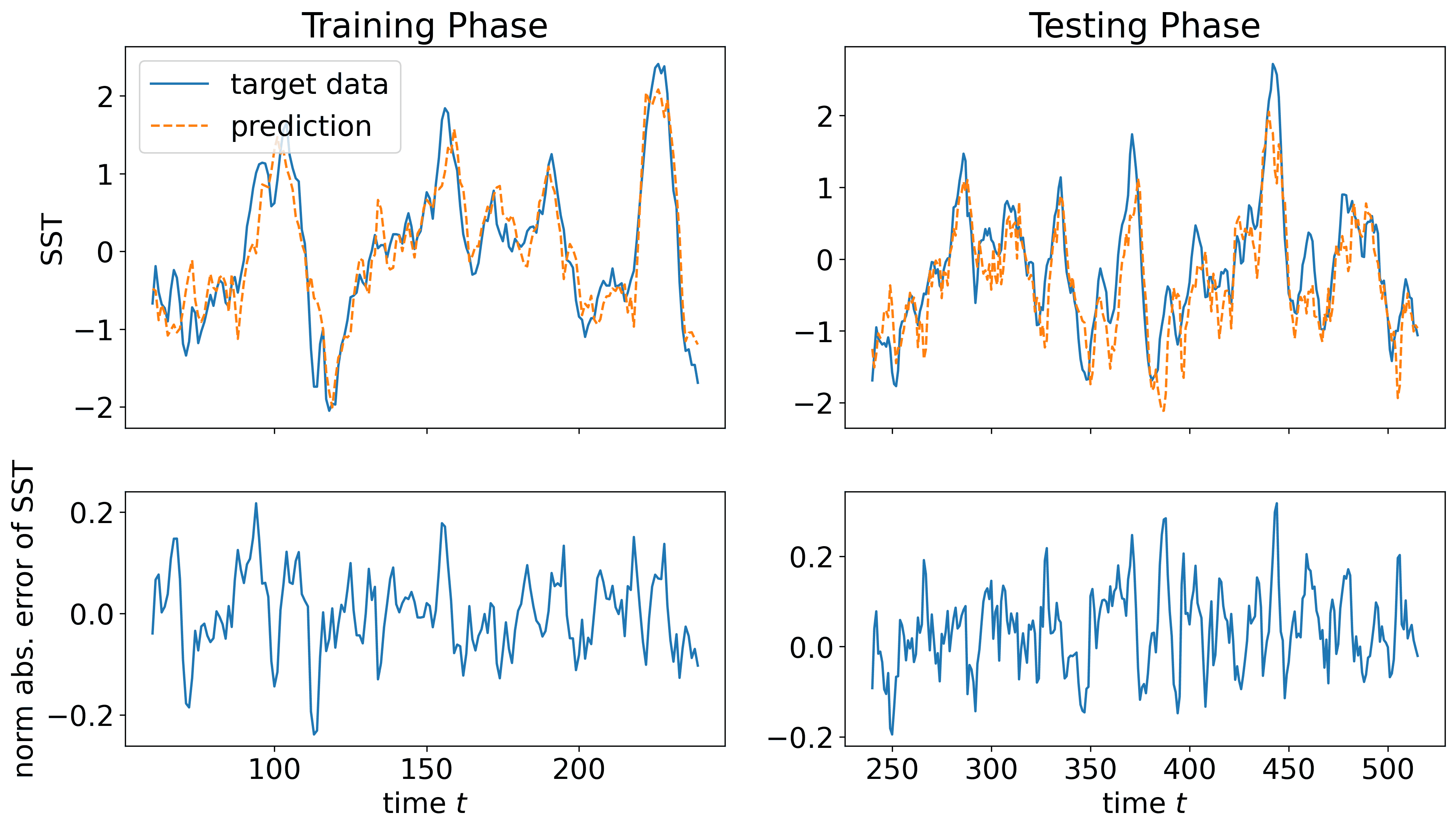}
		\caption{Optimal RC performance, $\text{NRMSE}_\text{RC} = 2.7 \times 10^{-1}$. Top panel: The target is shown in blue, the prediction with dashed orange. Bottom panel: normalized absolute error in each time step.}
	\end{subfigure}
	\caption{\label{fig:optplot_enso_sst}ENSO system inferring SST. $\Delta t = 1$ represents monthly time steps. The left side shows the training, the right side the testing phase which are the same for both algorithms.}
\end{figure}

\begin{figure}[!htp]
	\centering
	\begin{subfigure}{\textwidth}
		\centering
		\includegraphics[width=.8\textwidth]{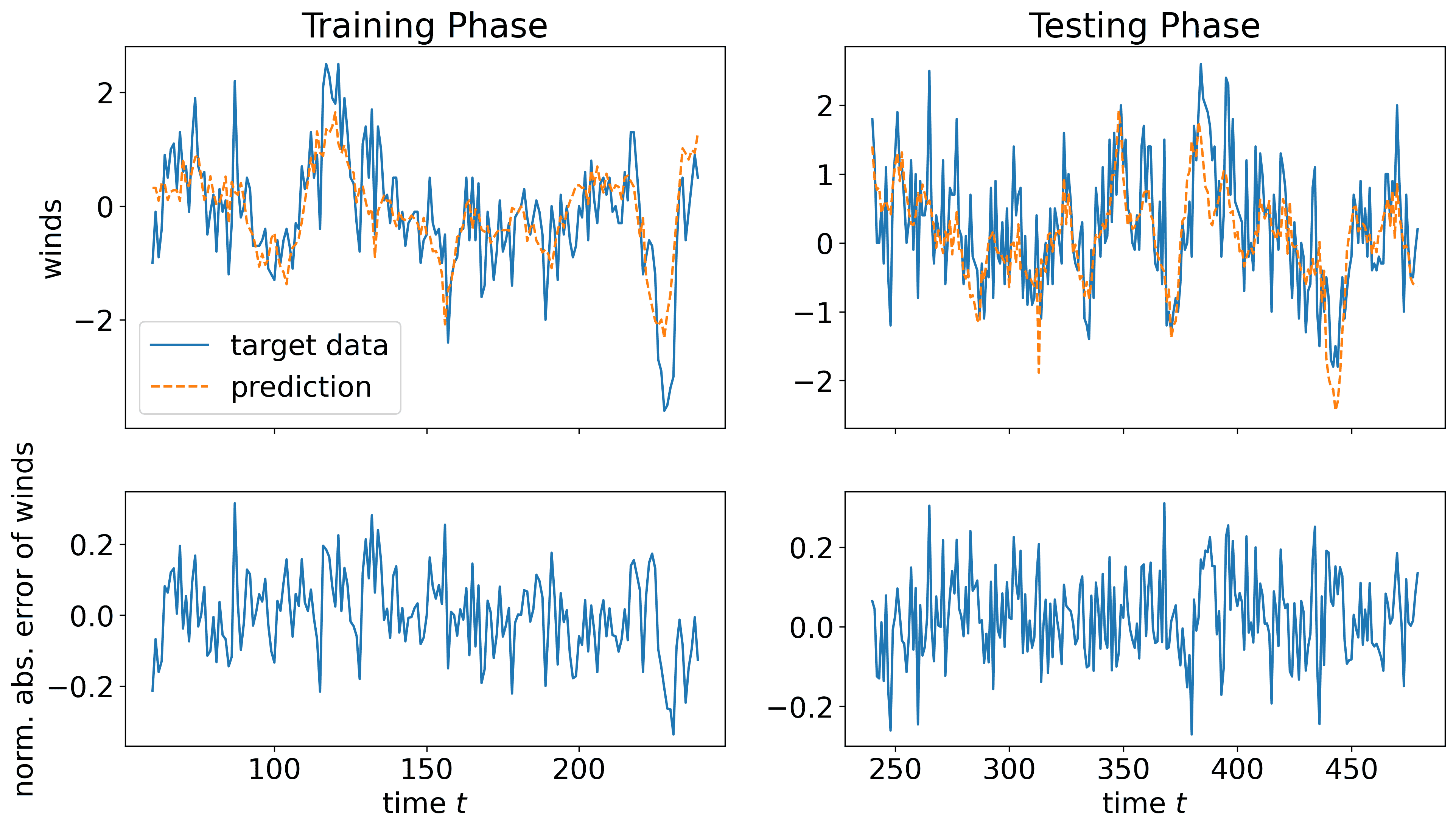}
		\caption{Optimal NGRC performance, $\text{NRMSE}_\text{NGRC} = 1.5 \times 10^{-1}$. Top panel: The target is shown in blue, the prediction with dashed orange. Bottom panel: normalized absolute error in each time step.}
	\end{subfigure}
	\vspace{0.5cm}
	\begin{subfigure}{\textwidth}
		\centering
		\includegraphics[width=.8\textwidth]{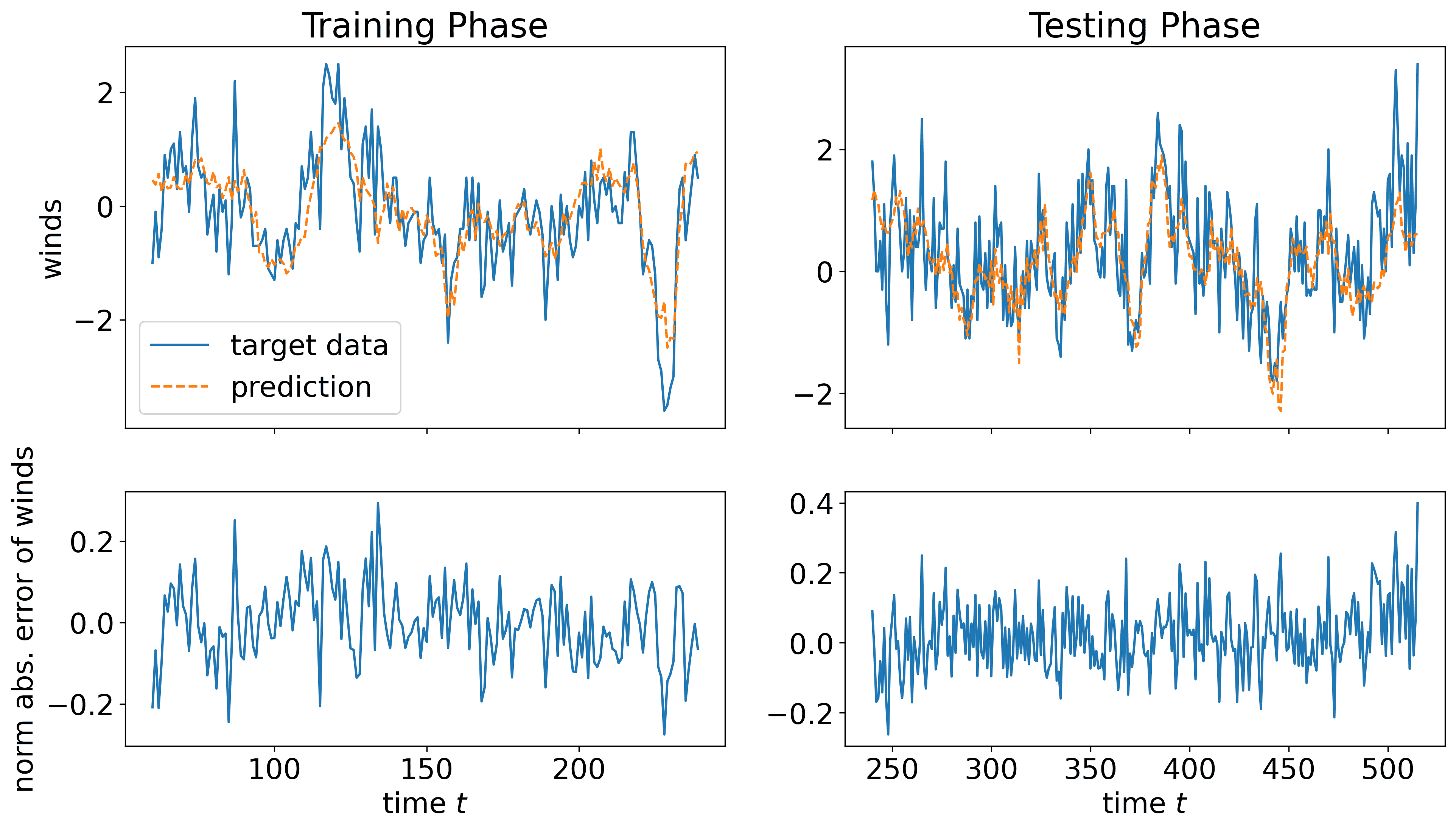}
		\caption{Optimal RC performance, $\text{NRMSE}_\text{RC} = 2.6 \times 10^{-1}$. Top panel: The target is shown in blue, the prediction with dashed orange. Bottom panel: normalized absolute error in each time step.}
	\end{subfigure}
	\caption{\label{fig:optplot_enso_winds}ENSO system inferring zonal winds. $\Delta t = 1$ represents monthly time steps. The left side shows the training, the right side the testing phase which are the same for both algorithms.}
\end{figure}

\end{document}